\documentstyle[epsfig,amsmath,proceed,times]{article} 

\newtheorem{theorem}{Theorem}

\newcommand{\mdp}{{\sc mdp}~}

\newcommand{\reinf}{{\sc reinforce}~}
\newcommand{\ipsg}{\mbox {\sc ipsg}}
\newcommand{\poipsg}{\mbox {\sc poipsg}}

\newcommand{\commentout}[1]{}
\newcommand{\tuple}[1]{\langle #1 \rangle}

\title{Learning to Cooperate via Policy Search}

\author{ {\bf Leonid Peshkin} \\ 
{\tt pesha@ai.mit.edu}\\
MIT AI Laboratory \\
545 Technology Square \\
Cambridge, MA 02139 \\
\And
{\bf Kee-Eung Kim} \\
{\tt kek@cs.brown.edu} \\
Computer Science Dept.  \\
Brown University, Box 1910 \\
Providence, RI 02906 \\
\And
{\bf Nicolas Meuleau} \\
{\tt nm@ai.mit.edu}\\
MIT AI Laboratory \\
545 Technology Square \\
Cambridge, MA 02139\\
\And
{\bf Leslie Pack Kaelbling} \\
{\tt lpk@ai.mit.edu} \\
MIT AI Laboratory \\
545 Technology Square \\
Cambridge, MA 02139\\
}

\begin{document}
\maketitle 

\begin{abstract}
Cooperative games are those in which both agents share the same payoff
structure. Value-based reinforcement-learning algorithms, such as variants of
Q-learning, have been applied to learning cooperative games, but they only
apply when the game state is completely observable to both agents. Policy
search methods are a reasonable alternative to value-based methods for
partially observable environments. In this paper, we provide a gradient-based
distributed policy-search method for cooperative games and compare the notion
of local optimum to that of Nash equilibrium. We demonstrate the
effectiveness of this method experimentally in a small, partially observable
simulated soccer domain.
\end{abstract}

\section{INTRODUCTION}
\label{intro}

The interaction of decision makers who share an environment is traditionally
studied in game theory and economics. The game 
theoretic formalism is very general, and analyzes the problem in terms of
solution concepts such as Nash equilibrium~\cite{Osborne}, but usually works
under the assumption that the environment is perfectly known to the agents.

In reinforcement learning~\cite{Kaelbling96,Sutton98a}, no explicit model of the
environment is assumed, and learning happens through trial and error.
Recently, there has been interest in applying reinforcement learning
algorithms to multi-agent environments. For example, Littman~\cite{Littman94}
describes and analyzes a {\it Q-learning}-like algorithm for finding optimal
policies in the framework of zero-sum Markov games, in which two players have
strictly opposite interests. Hu and Wellman~\cite{Wellman} propose a
different multi-agent Q-learning algorithm for {\it general-sum} games, and
argue that it converges to a Nash equilibrium.

A simpler, but still interesting case, is when multiple agents \\ share the same
objectives. A study of the behavior of agents employing Q-learning
individually was made by Claus and Boutilier~\cite{Boutilier97}, focusing on
the influence of game structure and exploration strategies on convergence to
Nash equilibria. In Boutilier's later work~\cite{BoutilierIJCAI99}, an extension
of value iteration was developed that allows each agent to reason explicitly
about the state of coordination.

However, all of this research assumes that the agents have the ability to
completely and reliably observe both the state of the environment and the
reward received by the whole system. 
Schneider et al.~\cite{Schneider99} investigate a case of distributed
reinforcement learning, in which agents have complete and reliable state
observation, but only receive a local reinforcement signal. They investigate
rules that allow individual agents to share reinforcement with their
neighbors.
In this paper we investigate the complementary problem in which the agents
all receive the shared reward signal, but have incomplete, unreliable, and
generally different perceptions of the world state. 
In such environments, value-search methods are generally inappropriate,
causing us to turn to policy-search methods~\cite{Williams87,Baird,Baird99}
which we have applied previously to single-agent partially observable
domains~\cite{MeuleauUAI99,Peshkin99}. 

In this paper we describe a gradient-descent policy-search algorithm for
cooperative multi-agent domains. In this setting, after each agent performs
its action given its observation according to some individual strategy, they
all receive the same payoff. Our objective is to find a learning algorithm
that makes each agent independently find a strategy that enables the group of
agents to receive the optimal payoff. Although this will not be possible in
general, we present a distributed algorithm that finds {\em local} optima in
the space of the agents' policies.

The rest of the paper is organized as follows. In section~\ref{IPG}, we give
a formal definition of a cooperative multi-agent environment. In
section~\ref{GD}, we review the gradient descent algorithm for policy search,
then develop it for the multi-agent setting. In section~\ref{BS}, we discuss
the different notions of optimality for strategies. Finally, we present
empirical results in section~\ref{experiments}.

\section{IDENTICAL PAYOFF GAMES}
\label{IPG}

An {\it identical payoff stochastic game}\footnote{\ipsg's are also
called {\em stochastic games}~\cite{Wellman}, Markov games~\cite{Littman94}
and {\em multi-agent Markov decision processes}~\cite{BoutilierIJCAI99}.}
(\ipsg)~\cite{Narendra} describes the 
interaction of a set of agents with a Markov environment in which they all
receive the same payoffs. An (\ipsg) is a tuple
$\tuple{\!S,\pi^S_0,G,T,r\!}$, where $S$ is a discrete state
space; $\pi^S_0$ is a probability distribution\footnote{Without loss of 
generality, we assume a degenerate distribution with one initial state
and omit it in our notation.} over the initial state; $G$ is
a collection of agents, where an {\it agent} $i$ is a 3-tuple,
$\tuple{\!A^i,O^i,B^i\!}$, of its discrete action space $A^i$,
discrete observation space $O^i$, and observation function \mbox{$B^i\!: S\!
\rightarrow\!{\cal P}(O^i)$};
$T\!:\!S\!\times\!{\cal A}\!\rightarrow\!{\cal P}(S)$ is a mapping from
states of the environment and actions of the agents to probability
distributions over states of the environment\footnote{${\cal P}(\Omega)$
denotes the set of probability distributions defined on some space $\Omega$.};
and \mbox{$r\!:\!S\!\times\!{\cal A} \!\rightarrow\!{\cal R}$} is the payoff
function, where ${\cal A} = \prod_i A^i$ is the joint action space of the
agents. 
When all agents in $G$ have the identity observation function $B(s)\!=\!s$ for
all $s\!\in\!S$, the game is {\it completely observable}. Otherwise, it is a
{\it partially observable} \ipsg~(\poipsg). 

In a \poipsg, at each time step: each agent \mbox{$i\!\in\!(1..m)$} observes
$o_i(t)$ corresponding to $B^i(s(t))$ and selects an action $a_i(t)$
according to its strategy; a compound action $\vec a(t)=(a_1(t), \ldots,
a_m(t))$ from the joint action space ${\cal A}$ is performed, inducing a
state transition of the environment; and the identical reward $r(t)$ is
received by all agents.
%; and each agent updates its policy.

The objective of each agent is to choose a strategy that maximizes the
{\it value of the game}. For a discount factor
\mbox{$\gamma\!\in\![0,1)$} and a set of strategies ${\vec
\mu}\!=\!(\mu_1,\ldots,\mu_m)$, the value of the game is
\begin{equation}
\label{value}
V(\vec \mu)=\sum_{t=0}^{\infty} \gamma^t {\rm E}_{\vec \mu} \left[r(t)\right]\;\;.
\end{equation}
In the general case, a {\it strategy} for some agent is a mapping from the
history of all observations from the beginning of the game into the current
action $a(t)$. We limit our consideration in this paper to cases in which the
agent's actions may depend only on the current observation, or in which the
agent has a finite internal memory. When actions depend only on the current
observation, the policy is called a {\it memoryless} or {\it reactive
policy}. When this dependence is probabilistic, we call it a {\it stochastic
reactive policy}, otherwise a {\it deterministic reactive policy}.

\begin{figure}[ht]
\centerline{\epsfig{file=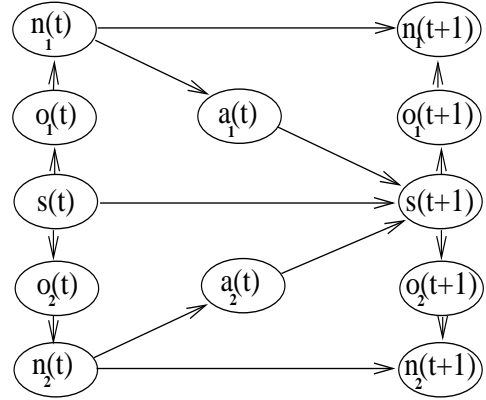,width=2.5in,height=2.1in}}
\caption{An influence diagram for two agents with {\sc fsc}s in a {\sc poipsg}.}
\label{mapg}
\end{figure}

Note that in a completely observable \ipsg, reactive policies are sufficient
to implement the best possible joint strategy. This follows directly from the
fact that every \mdp has an optimal deterministic reactive
policy~\cite{Puterman94}. Therefore an \mdp with the product action space
$\prod_i A^i$ corresponding to a completely observable \ipsg~ also has one,
representable by deterministic reactive policies for each agent. However, it
has been shown that in partially observable environments, the best reactive
policy can be arbitrarily worse than the best policy using
memory~\cite{Singh94a}. This statement can also be easily extended to {\sc
poipsg}s.

There are many possibilities for constructing policies with 
memory~\cite{Peshkin99,MeuleauUAI99}. In this work we use a {\it finite state
controller} ({\sc fsc}) for each agent. A more detailed description of {\sc
fsc}s and derivation of algorithms for learning them may be found in a
previous paper~\cite{MeuleauUAI99}; we simply state the definition here.

{\it A finite state controller} ({\sc fsc}) for an agent with action space
$A$ and observation space $O$ is a tuple $\tuple{\!N,\pi^N_0,\eta,\psi}$, 
where $N$ is a finite set of
internal controller states, $\pi^N_0$ is a probability distribution over the
initial internal state, $\eta\!:\!N\!\times\!O\!\rightarrow\!
{\cal P}(N)$ is the internal state transition function that maps an internal
state and observation into a probability distribution over internal
states, and $\psi:\!N\!\rightarrow\!{\cal P}(A)$ is the action
function that maps an internal state into a probability
distribution over actions. Figure~\ref{mapg} depicts an influence diagram for
two agents controlled by {\sc fsc}s.

Note that in partially observable environments, agents controlled by {\sc
fsc}s might not have enough memory to even represent an optimal policy which
could, in general, require infinite memory, as in a partially observable
Markov decision process ({\sc pomdp})~\cite{Sondik78}. In this paper, we
concentrate on the problem of finding the (locally) optimal controller from
the class of {\sc fsc}s with some fixed size of memory. 

\begin{figure}[th]
\centerline{\epsfig{file=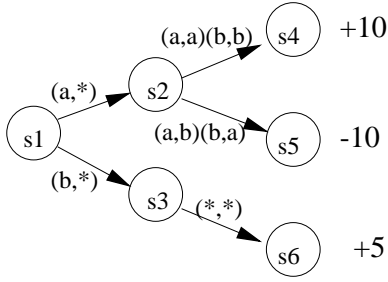,width=2.0in}}
\caption{A coordination problem in a completely observable identical payoff game.}
\label{craig}
\end{figure}

To better understand {\sc ipsg}s, let us consider an example from
Boutilier~\cite{BoutilierIJCAI99}, illustrated in figure~\ref{craig}. There are
two agents, $a_1$ and $a_2$, each of which has a choice of two
actions, $a$ and $b$, at any of three states. All transitions are deterministic
and are labeled by the joint action that corresponds to the transition. For
instance, the joint action $(a,b)$ corresponds to the first agent performing
action $a$ and the second agent performing action $b$. Here, $*$ refers to
any action taken by the corresponding agent.

The starting state is $s1$, where the first agent alone decides whether to
move the environment to state $s2$ by performing action $a$ or to state $s3$
by performing action $b$. In state $s3$, no matter what both agents do as the
next step, they receive a reward of $+5$ in state $s6$ risk-free. In state
$s2$, the agents have a choice of cooperating---choosing the same action,
whether $(a,a)$ or $(b,b)$---with reward $+10$ in state $s4$, or
not---choosing different actions, whether $(a,b)$ or $(b,a)$---and
getting $-10$ in state $s5$.

We will represent a joint policy with parameters $p^{\rm Agent}_{\rm State}$,
denoting the probability that an agent will perform action $a$ in the
corresponding state. Only three parameters are important for the outcome:
$\{p^1_1,p^1_2;p^2_2\}$. The optimal joint policies are $\{1,1;1\}$ or
$\{1,0;0\}$, which are deterministic reactive policies.

\section{GRADIENT DESCENT FOR POLICY SEARCH}
\label{GD}

In this section, we first introduce a general method for using gradient
descent in policy spaces, then show how it can be applied to multi-agent
problems.

\subsection{Basic Algorithm}
\label{SC}

Williams introduced the notion of policy search for reinforcement learning in
his \reinf~ algorithm~\cite{Williams87,Williams92}, which was generalized to
a broader class of error criteria by Baird and Moore~\cite{Baird,Baird99}.

We will start by considering the case of a single agent interacting with
a {\sc pomdp}. The agent's policy $\mu$ is assumed to depend on some internal
state taking on values in finite set $N$. We will not make any further
commitment to details of the policy's architecture, as long as it defines
the probability of action given past history as a continuous differentiable
function of some set of parameters $w$.

First, we will establish some notation. We denote by $H_t$ the set of all
possible experience sequences of length $t$: $h=\tuple{\!n(0)\!,o(1)\!,
n(1)\!,a(1)\!,r(1)\!,\ldots\!,o(t)\!,n(t)\!,a(t)\!,r(t)\!,o(t\!+\!1)\!}$. 
In order to
specify that some element is a part of the history $h$ at time $\tau$, we
write, for example, $r(\tau,\!h)$ and $a(\tau,h)$ for the $\tau^{th}$ reward
and action in the history $h$. We will also use $h^\tau$ to denote a prefix
of the sequence $h\!\in\!H_t$ truncated at time $\tau\!\leq\!t\!:$
$h^\tau\!\stackrel{\rm def}{=}\!\langle n(0)\!,o(1)\!,n(1),$
$a(1),r(1),\ldots,o(\tau),n(\tau),a(\tau),r(\tau),o(\tau\!+\!1)\rangle$.
The value defined by equation~(\ref{value}) can be rewritten as
\begin{equation}
\label{eqB}
V(\mu)=\sum_{t=1}^\infty \gamma^t \sum_{h \in H_t} \Pr(h \mid \mu) r(t,h)\;\;.
\end{equation}

Let us assume the policy is expressed parametrically in terms of a vector of
weights $\vec w\!=\!\{w_1,\ldots,w_M \}$. If we could calculate the
derivative of $V$ for each $w_k$, it would be possible to do an exact
gradient descent on value $V$ by making updates $ \Delta w_k = \alpha
\frac{\partial}{\partial w_k}V.$ We can compute the derivative for each
weight $w_k$,
\begin{equation*}
\begin{split}
\frac{\partial V(\mu)}{\partial w_k}\!=\!&\sum_{t =
1}^\infty\!\gamma^t\!\sum_{h \in H_t}\!\left[r(t,h)\frac{\partial 
\Pr(h\!\mid\!\mu)}{\partial w_k}\right] \\ 
=&\sum_{t = 1}^\infty \gamma^t \sum_{h \in H_t} \Pr(h \mid \mu) r(t,h) \\
\times&\!\sum_{\tau=1}^t \frac{\partial \ln \Pr\left(n(\tau,h), 
a(\tau,h)\!\mid\!h^{\tau-1}, \mu \right)} {\partial w_k}\;\;.  
\end{split} 
\end{equation*}

But, in the spirit of reinforcement learning, we cannot assume the knowledge
of a world model that would allow us to calculate $\Pr(h\!\mid\!\mu)$,
so we must retreat to stochastic gradient descent instead. We sample from the
distribution of histories by interacting with the environment, and calculate
during each trial an estimate of the gradient, accumulating the quantities:
\begin{equation}
\label{eqE}
\gamma^t r(t,h) \sum_{\tau=1}^t \frac{\partial 
\ln \Pr\left(n(\tau,h), a(\tau,h) \mid h^{\tau-1}, \mu \right)}
{\partial w_k}\;\;,  
\end{equation} 
%\frac{\partial V}{\partial w_k} \!\approx\!
for all $t$. For a particular policy architecture, this can be readily
translated into a gradient descent algorithm that is guaranteed to converge
to a local optimum of $V$.

\subsection{Central Control Of Factored Actions}
\label{cent}

Now let us consider the case in which the action is factored, meaning that
each action $\vec a$ consists of several components $\vec
a\!=\!(a_1,\ldots,a_m)$. We can consider two kinds of control\-lers: a {\em
joint controller} is a policy mapping observations to the complete joint
distribution $\pi (\vec a)$; a {\em factored controller} is made up of
independent sub-policies $\mu_{a_i}\!:\!O^i\!\rightarrow\!{\cal P}(a_i)$
(possibly with a dependence on individual internal state) for each action
component.

Factored controllers can represent only a subset of the policies represented
by joint controllers. Obviously, any product of policies for the factored
controller $\prod_i \mu_{a_i}$ can be represented by a joint controller
$\mu_{\vec a}$, for which $\Pr(\vec a) = \prod_{i=1}^N \Pr(a_i) $. However,
there are some stochastic joint controllers that cannot be represented by any
factored controller, because they require coordination of probabilistic
choice across action components, which we illustrate by the following example.

The first action component controls the liquid component of a meal
$a_1\!\in\!\{{\rm vodka,milk}\}$ and the second controls the solid one
$a_2\!\in\!\{{\rm pickles,cereal}\}$. For the sake of argument, let us assume
that sticking to one combination or another is not as good as a ``mixed
strategy'', meaning that for a healthy diet, we sometimes want to eat milk
with cereal, other times vodka with pickles. The optimal policy is
randomized, say $10\%$ of the time $\vec a\!=\!(vodka,pickles)$ and $90\%$ of
the time $\vec a\!=\!(milk,cereal)$. But when the components are controlled
independently, we cannot represent this policy. With randomization, we are
forced to drink vodka with cereal or milk with pickles on some occasions.

Because we are interested in individual agents learning independent policies,
we concentrate on learning the best factored controller for a domain, even if
it is suboptimal in a global sense. Requiring a controller to be factored
simply puts constraints on the class of policies, and therefore distributions
$P(a\!\mid\!\mu, h)$, that can be represented. The stochastic gradient-descent
techniques of the previous section can still be applied directly in this case
to find local optima in the controller space. We will call this method {\it
joint gradient descent}.

\subsection{Distributed Control Of Factored Actions}
\label{MA}

The next step is to learn to choose action components not centrally, but
under the distributed control of multiple agents. One obvious strategy
would be to have each agent perform the same gradient-descent algorithm in
parallel to adapt the parameters of its own local policy $\mu_{a_i}$. Perhaps
surprisingly, this {\it distributed gradient descent} ({\sc dgd}) method is
very effective.

\begin{theorem}
For factored controllers, distributed gradient descent is equivalent to 
joint gradient descent.
\end{theorem}
{\em Proof:} We will show that for both controllers the algorithm will
be stepwise the same, so starting from the same point in the search
space, on the same data sequence, the algorithms will converge to the
same locally optimal parameter setting. For simplicity we assume a
degenerate set of internal controller states $N$. Then 
a factored controller, $\vec h$ can be described as 
$\langle o_1(1),...,o_m(1),a_1(1),...,a_m(1),r(1),...\rangle$. 
The corresponding history for an individual agent $i$ is
${h_i}=\langle o_i(1),a_i(1),r(1),...\rangle$. It is clear
that a collection $h_1...h_m$ of individual histories, one for each agent,
specifies the joint history $\vec h$.

The joint gradient descent algorithm requires that we draw sample histories
from $\Pr(\vec h\!\mid\!\vec \mu)$ and that we do gradient descent
on $\vec w$ with a sample of the gradient at each time step $t$ in the
history equal to 
$$
\gamma^t r(t,h) \sum_{\tau=1}^t \frac
{\partial\ln \Pr(a(\tau,h) \mid \vec h^{\tau-1},\mu)}{\partial w}\;.
$$

Whether a factored controller is being executed by a single agent, or it is
implemented by agents individually executing policies $\mu_{a_i}$ in
parallel, joint histories are generated from the same distribution 
$\Pr(\vec h\!\mid\!\tuple{\mu_{a_1}\!,...,\mu_{a_m}}\!)$. So
the distributed algorithm is sampling from the correct distribution.

Now, we must show that the weight updates are the same in the distributed
algorithm as in the joint one. Let $\vec w_k\!=\!(w^0_k,\ldots,w^{M_k}_k)$ be
the set of parameters controlling action component $a_k$. Then
$$\frac{\partial}{\partial w^j_k} \ln \left(\Pr\left(a_l(\tau) \mid 
\vec h_l^{\tau-1},\mu_l\right)\right)=0\;\; {\rm for\; all}\;\; k \neq l\;;
$$ that is, the action probabilities of agent $l$ are independent of the 
parameters in other agents' policies. With this in mind, for factored 
controllers, the derivative in expression~(\ref{eqE}) becomes
\begin{equation*}
\begin{split}
\frac{\partial}{\partial w^j_k} \ln & \Pr\left(\vec a(\tau,h) \mid 
     \vec h^{\tau-1},\vec \mu\right) \\
&=\frac{\partial}{\partial w^j_k} \ln \prod_{i=1}^m \Pr\left(a_i(\tau,h) 
    \mid h_i^{\tau-1},\mu_i\right)\\
&=\sum_{i=1}^m \frac{\partial}{\partial w^j_k} \ln \Pr\left(a_i(\tau,h) 
   \mid h_i^{\tau-1}, \mu_i\right)\\
& = \frac{\partial}{\partial w^j_k} \ln \Pr\left(a_k(\tau,h) 
    \mid h_k^{\tau-1}, \mu_k\right)\;.
\end{split}
\end{equation*}
Thus, the same weight updates will be performed by {\sc dgd} as by joint
gradient descent on a factored controller. \rule{1.8mm}{1.8mm} 

This theorem shows that policy learning and control over component actions
can be distributed among independent agents who are not aware of each others'
choice of actions. An important requirement, though, is that agents perform
simultaneous learning (which might be naturally synchronized by the coming of
the rewards).
%, and keep the same learning rates at all times during the learning period.

\section{RELATING LOCAL OPTIMA TO NASH EQUILIBRIA}
\label{BS}

In game theory, the Nash equilibrium is a common solution concept. Because
gradient descent methods can often be guaranteed to converge to local optima
in the policy space, it is useful to understand how those points are related
to Nash equilibria. We will limit our discussion to the two-agent case, but
the results are generalizable to more agents.

A Nash equilibrium is a pair of strategies such that deviation by one agent
from its strategy, assuming the other agent's strategy is fixed, cannot
improve the overall performance. Formally, in an \ipsg, a {\it Nash
equilibrium} point is a pair of strategies $(\mu_1^*,\mu_2^*)$ such that:
\begin{equation*}
\begin{split}
V\left(\tuple{\mu_1^*, \mu_2^*}\right) &\geq 
    V\left(\tuple{\mu_1, \mu_2^*}\right), \\ 
V(\tuple{\mu_1^*, \mu_2^*}) &\geq V(\tuple{\mu_1^*, \mu_2}) \;  
\end{split}
\end{equation*}
for all $\tuple{\mu_1, \mu_2}$. When the inequalities are strict, it is
called a {\it strict Nash equilibrium}. 

Every discounted stochastic game has at least one Nash equi\-librium
point~\cite{Wellman}. It has been shown that under certain convexity
assumptions about the shape of payoff functions, the gradient-descent process
converges to an equilibrium point~\cite{Arrow60}. 
%For the case of \ipsg~ for example it would require that payoff function is
%concave in all policy parameters at the point of equilibrium.
It is clear that the optimal Nash equi\-librium point (the Nash equi\-librium
with the highest value) in an \ipsg~ also is a possible point of convergence
for the gradient descent algorithm, since it is the global optimum in the
policy space.

Let us return to the game described in Figure~\ref{craig}. It has two \\
optimal strict Nash equilibria at $\{1,1;1\}$ and $\{1,0;0\}$. It also has a
set of sub-optimal Nash equilibria $\{0,p^1_2;p^2_2\}$, where $p^2_2$ can
take on any value in the interval $[.25,.75]$ and $p^1_2$ can take any value
in the interval $[0,1]$. The sub-optimal Nash equilibria represent situations
in which the first agent always chooses the bottom branch and the second
agent acts moderately randomly in state $s2$. In such cases, it is
strictly better for the first agent to stay on the bottom branch with
expected value $+5$. For the second agent, the payoff is $+5$ no matter how
it behaves, so it has no incentive to commit to a particular action in state
$s2$ (which is necessary for the upper branch to be preferred).

In this problem, the Nash equilibria are also all local optima for the
gradient descent algorithm. Unfortunately, this equivalence only holds in one
direction in the general case. We state this more precisely in the following
theorems. 

\begin{theorem}
Every strict Nash equilibrium is a local optimum for gradient descent in the
space of parameters of a factored controller. 
\end{theorem}
{\em Proof:} Assume that we have two agents and denote the stra\-te\-gy at the
point of strict Nash equilibrium as $(\mu^*_1, \mu^*_2)$ encoded by parameter
vector $\tuple{w^1_1 \ldots w_1^G, w_2^1 \ldots w_2^G}$. For
simplicity, let us further assume that $(\mu^*_1, \mu^*_2)$ is not on the
boundary of the parameter space, and each weight is locally relevant: that is,
that if the weight changes, the policy changes, too.

By the definition of Nash equilibrium, any change in value of the parameters
of one agent without change in the other agent's parameters results in a
decrease in the value $V$. In other words, we have that $\partial V/\partial
w^j_i\!\le\!0$ \textit{and} $-\partial V/\partial w^j_i\!\le\!0$ for all $j$
and $i$ at the equilibrium point. Thus, $\partial V/\partial w^j_i = 0$ for
all $w^j_i$ at $(\mu^*_1, \mu^*_2)$, which implies it is a singular point of
$V$. Furthermore, because the value decreases in every direction, it must be
a maximum.

In the case of a locally irrelevant parameter $w^j_i$, $V$ will have a
ridge along its direction.  All points on the ridge are singular and,
although they are not strict local optima, they are essentially local
optima for gradient descent. \rule{1.8mm}{1.8mm}

The problem of Nash equilibria on the boundary of the parameter space is an
interesting one. Whether or not they are convergence points depends on the
details of the method used to keep gradient descent within the boundary. A
particular problem comes up when the equilibrium point occurs when one or
more parameters have infinite value (this is not uncommon, as we shall see in
section~\ref{experiments}). In such cases, the equilibrium cannot be reached,
but it can usually be approached closely enough for practical purposes.

\begin{theorem}
Some local optima for gradient descent in the space of parameters of a
factored controller are not Nash equilibria.
\end{theorem}
{\em Proof:} Consider a situation in which each agent's policy has a single
parameter $w_i$, so the policy space can be described by $\tuple{\!w_1,
w_2\!}$. We can construct a value function $V(w_1,w_2)$ such that for
some $c$, $V(\cdot,c)$ has two modes, one at $V(a,c)$ and the other at
$V(b,c)$, such that $V(b,c)\!>\!V(a,c)$. Further assume that $V(a,\cdot)$  and
$V(b,\cdot)$ each have global maxima $V(a,c)$ and $V(b,c)$. Then $V(a,c)$ is
a local optimum that is not a Nash equilibrium. \rule{1.8mm}{1.8mm}

%Note that if the statement of this theorem were false we could, by
%constructing a degenerate game with one player, prove that \reinf always
%converges to a global optimum.

\section{EXPERIMENTS}
\label{experiments}

There are no established benchmark problems for multi-agent learning. To
illustrate our method we present empirical results for two problems: the
simple coordination problem of figure~\ref{craig} and a small multi-agent
soccer domain.

\subsection{Simple Coordination Problem}
\label{Craig}

\begin{figure}[th]
\centerline{\epsfig{file=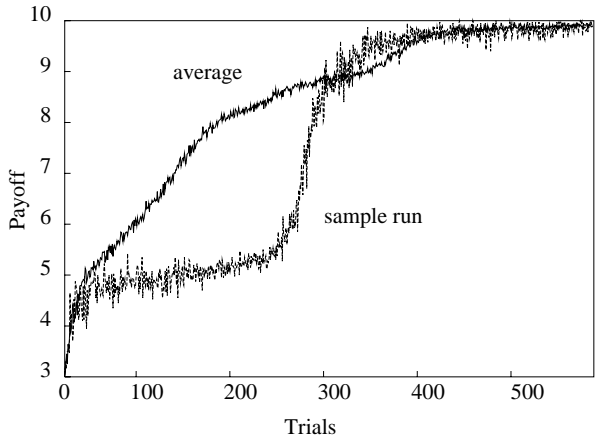,width=3.5in}}
\caption{Average payoff and payoff of a sample run of a distributed gradient
descent on the simple problem.} 
\label{empir1}
\end{figure}

We originally discussed policies for this problem using three weights,
one corresponding to each of the $p^{\rm Agent}_{\rm State}$
probabilities.  However, to force gradient descent to respect the
bounds of the simplex, we used the standard Boltzmann encoding, so
that for agent $i$ in state $s$ there are two weights, $w^i_{s,a}$ and
$w^i_{s,b}$, one for each action. The action probability is coded as a
function of these weights as
$$
p_s^i = \frac{e^{w^i_{sa}/\theta}}{e^{w^i_{sa}/\theta}+e^{w^i_{sb}/\theta}}>0.
$$

We ran {\sc dgd} with a learning rate of $\alpha\!=\!.003$ and a discount
factor of $\gamma\!=\!.99$; the results are shown in figure~\ref{empir1}. The
graph of a sample run illustrates how the agents typically initially move
towards a sub-optimal policy. The policy in which the first agent always
takes action $b$ and the second agent acts fairly randomly is a Nash
equilibrium, as we saw in section~\ref{BS}. However, this policy is not
exactly representable in the Boltzmann parameterization because it requires
one of the weights to be infinite to drive a probability to either $0$ or
$1$.

So, although the algorithm moves toward this policy, it never reaches
it exactly. This means that there is an advantage for the second agent to
drive its parameter toward $0$ or $1$, resulting in eventual convergence
toward a global optimum (note that, in this parameterization, these optima
cannot be reached exactly, either). The average of $10$ runs shows that the
algorithm always converges to a pair of policies with value very close to the
maximum value of $10$.

\subsection{Soccer}
\label{soccer}

\begin{figure}[th]
\vspace{.3in}
\centerline{\epsfig{file=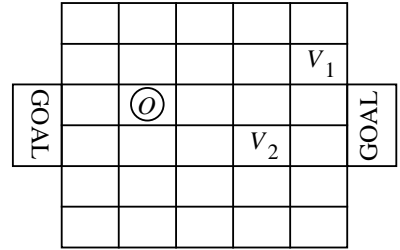,height=1.30in}}
\caption{The soccer field. $V_1$ and $V_2$ represent learning agents 
and $O$ represents the opponent.}
\label{fig:soccer_field}
\end{figure}

We have also conducted experiments on a small soccer domain adapted from
Littman~\cite{Littman94}. The game is played on a $6\!\times\!5$ grid as shown in
Figure~\ref{fig:soccer_field}. There are two learning agents on one team and
a single opponent with a fixed strategy on the other. Every time the game
begins, the learning agents are randomly placed in the right half of the
field, and the opponent in the left half of the field. Each cell in the grid
contains at most one player. Every player on the field (including the
opponent) has an equal chance of initially possessing the ball.

\begin{figure*}[th]
\epsfig{file=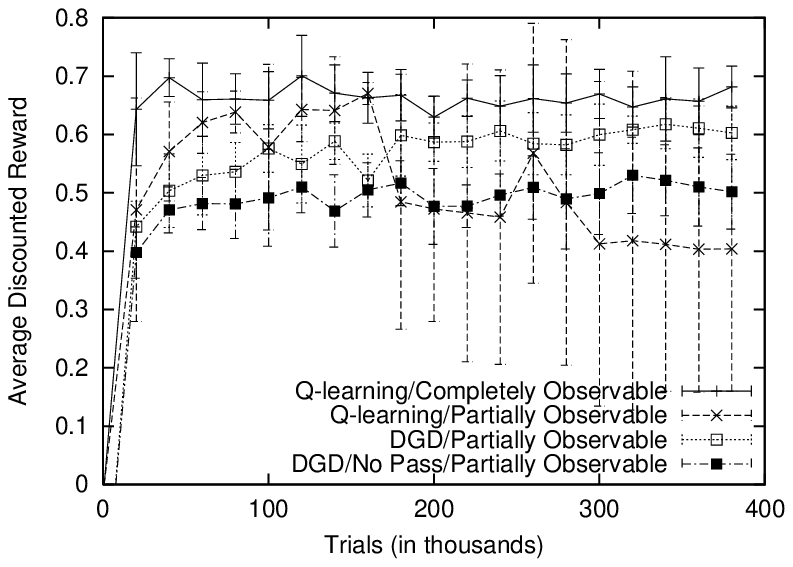,width=3.2in}\epsfig{file=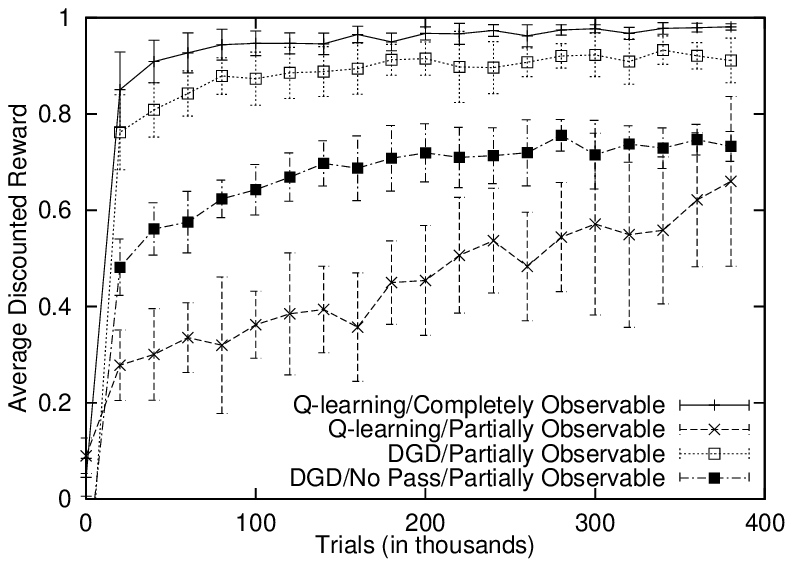,width=3.2in}
\epsfig{file=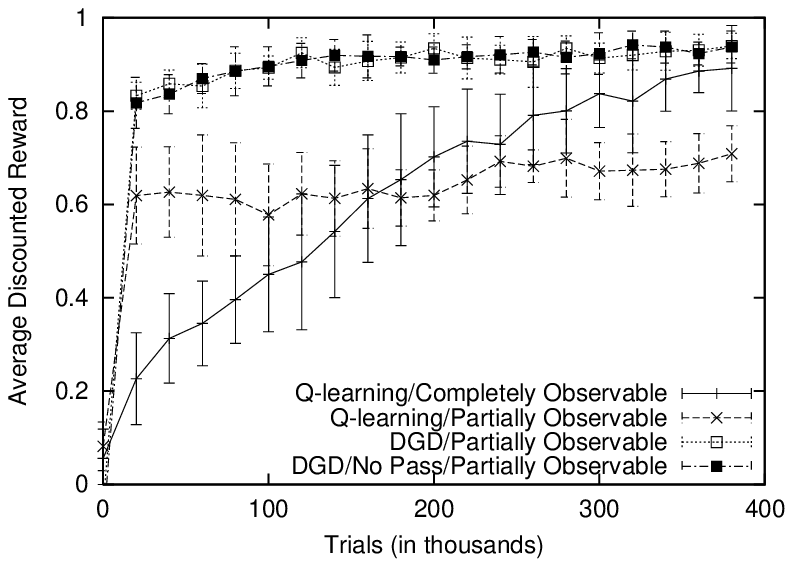,width=3.2in}\epsfig{file=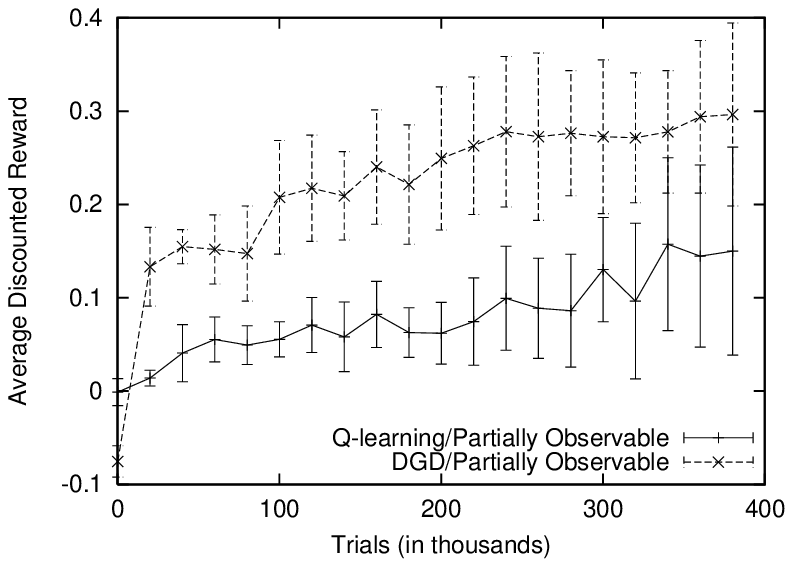,width=3.2in}
\caption{Learning curves of {\sc dgd} (policy search) and Q-learning against
a defensive opponent (top left), a greedy opponent (top right), a random opponent
(bottom left), and a team with two agents with different fixed strategies
(bottom right).}
\label{fig:soccer_graphs}
\end{figure*}

At each time step, a player can execute one of the six actions:
$\{North,South,East,West, Stay, Pass\}$. When an agent passes, the ball is
transferred to the other agent on its team on the next time step. Once all
players have selected actions, they are executed in a random order. When a
player executes an action that would move it into the cell occupied by some
other player, possession of the ball goes to the stationary player and the
move does not occur. When the ball falls into one of the goals, the game ends
and a reward of $\pm 1$ is given. 

We made a partially observable version of the domain to test the
effectiveness of {\sc dgd}: each agent can only obtain information about which
player possesses the ball and the status of cells to the north, south, east
and west of its location. There are 3 possible observations for each cell:
whether it is open, out of the field, or occupied by someone. In
Figure~\ref{fig:soccer_graphs}, we compare the learning curves of {\sc dgd}
to those of Q-learning with a central controller for both the completely
observable and the partially observable cases. We also show learning curves
of {\sc dgd} without the action \textit{Pass} in order to measure the
\textit{cooperativeness} of the learned policies.

The graphs in the figure summarize simulations of the game against three
different fixed-strategy opponents: \\ $\bullet$ Random: Executes actions
uniformly at random.\\ $\bullet$ Greedy: Moves toward the player possessing
the ball and stays there. Whenever it has the ball, it rushes to the goal.\\
$\bullet$ Defensive: Rushes to the front of its own goal, and stays or moves
at random, but never leaves the goal area.

We show the average performance over $10$ runs with error bars for standard
deviation. The learning rate was $0.05$ for {\sc dgd} and $0.1$ for
Q-learning, and the discount factor was $0.999$, throughout the experiments.
Each agent in the {\sc dgd} team learned a reactive policy. The policy's
parameters were initialized by drawing uniformly at random from the
appropriate domains. We used $\epsilon$-greedy exploration with
$\epsilon\!=\!0.4$ for Q-learning. The performance in the graph is reported
by evaluating the greedy policy derived from the Q-table.

Because, in the completely observable case, this domain is an \mdp (the
opponent's strategy is fixed, so it is not really an adversarial game),
Q-learning can be expected to learn the optimal joint policy, which it seems
to do. It is interesting to note the slow convergence of completely
observable Q-learning against the random opponent. We conjecture that this is
because, against a random opponent, a much larger part of the state space is
visited. The table-based value function offers no opportunity for
generalization, so it requires a great deal of experience to converge.

As soon as observability is restricted, Q-learning no longer reliably
converges to the best strategy. The joint Q-learner now has as its input the
two local observations of the individual players. It behaves quite
erratically, with extremely high variance because it sometimes converges to a
good policy and sometimes to a bad one. This unreliable behavior can be
attributed to the well-known problems of using value-function approaches, and
especially Q-learning, in {\sc pomdp}s.

The individual {\sc dgd} agents have stochasticity in their action choices,
which gives them some representational abilities unavailable to the
Q-learner. We tried additional experiments in which the {\sc dgd} agents had
4-state {\sc fsc}'s. Their performance did not improve appreciably. We expect
that, in future experiments on a larger field with more players, it will be
important for the agents to have internal state. 

Despite the fact that they learn independently, the combination of policy
search plus a different policy class allows them to gain considerably
improved performance. We cannot tell how close this performance is to the
optimal performance with partial observability, because it would be
computationally impractical to solve the \poipsg~ exactly. Bernstein et.
al.~\cite{BernsteinUAI00} show that in the finite-horizon case two-agent {\sc
poipsg}s are {\em harder} to solve than {\sc pomdp}s (in the worst-case
complexity sense).

It is also important to see that the two {\sc dgd} agents have learned to
cooperate in some sense: when the same algorithm is run in a domain without
the ``pass'' action, which allows one agent to give the ball to its teammate,
performance deteriorates significantly against both defensive and greedy
opponents. Since the agents don't know where they are with respect to the
goal, they probably choose to pass whenever they are faced with the opponent.
Against a completely random opponent, both strategies do equally well. It is
probably sufficient, in this case, to simply run straight for the goal, so
cooperation is not necessary.

We performed some additional experiments in a two-on-two domain in which one
opponent behaved greedily and the other defensively. In this domain, the
completely observable state space is so large that it is difficult to even
store the Q table, let alone populate it with reasonable values. Thus, we
just compare two 4-state {\sc dgd} agents with a limited-view centrally
controlled Q-learning algorithm. Not surprisingly, we find that the {\sc dgd}
agents are considerably more successful.
%, as shown in the learning curves in figure~\ref{fig:soccer_graphs}.

Finally, we performed informal experiments with an increasing number of
opponents. The opponent team was made up of one defensive agent and an
increasing number of greedy agents. For all cases in which the opponent team
had more than two greedy agents, {\sc dgd} led to a defensive strategy in
which, most of the time, the agents all rushed to the front of their goal and
stayed there forever.

\section{CONCLUSIONS AND FUTURE WORK}
\label{discussion}

We have presented an algorithm for distributed learning in cooperative
multi-agent domains.  It is guaranteed to find local optima in the
space of factored policies.  We cannot show, however, that it always
converges to a Nash equilibrium, because there are local optima in
policy space that are not Nash equilibria.  The algorithm performed
well in a small simulated soccer domain.

It will be important to apply this algorithm in more complex domains, to see
if the gradient remains strong enough to drive the search effectively, and to
see whether local optima become problematic. An interesting extension of this
work would be to allow the agents to perform explicit communication actions
with one another to see if they are exploited to improve performance in the
domain.

In addition, there may be more interesting connections to establish with game
theory, especially in relation to solution concepts other than Nash
equilibrium, which may be more appropriate in cooperative games.

%\commentout{economic theory, v13, p365-75, March99    99256}

{\footnotesize
\subsubsection*{Acknowledgement}
Craig Boutilier and Michael Schwarz gave relevant ideas and comments.
Franjo Ivanci\'c pointed out an inaccuracy in the description of the
basic algorithm.  Leonid Peshkin was supported by grants from NSF and
NTT; Nicolas Meuleau in part by research grant from NTT; Kee-Eung Kim
in part by AFOSR/RLF 30602-95-1-0020; and Leslie Pack Kaelbling in
part by a grant from NTT and in part by DARPA Contract \#DABT
63-99-1-0012.

\setlength{\baselineskip}{1mm}

}

\begin{thebibliography}{10}

\bibitem{Arrow60}
K.~J. Arrow and L.~Hurwicz.
\newblock Stability of the gradient process in $n$-person games.
\newblock {\em Journal of the Society for Industrial and Applied Mathematics},
  8(2):280--295, 1960.

\bibitem{Baird}
L.~C. Baird.
\newblock {\em Reinforcement Learning Through Gradient Descent}.
\newblock PhD thesis, Carnegie Mellon University, Pittsburgh, PA 15213, 1999.

\bibitem{Baird99}
L.~C. Baird and A.~W. Moore.
\newblock Gradient descent for general reinforcement learning.
\newblock In {\em Advances in Neural Information Processing Systems 11}. The
  MIT Press, 1999.

\bibitem{BernsteinUAI00}
D.~Bernstein, S.~Zilberstein, and N.~Immerman.
\newblock The compplexity of decentralized control of {Markov} decision
  processes.
\newblock In {\em Proceedings of the Sixteenth Conference on Uncertainty in
  Artificial Intelligence}, 2000.

\bibitem{BoutilierIJCAI99}
C.~Boutilier.
\newblock Sequential optimality and coordination in multiagent systems.
\newblock In {\em Proceedings of the Sixteenth International Joint Conference
  on Artificial Intelligence}, 1999.

\bibitem{Boutilier97}
C.~Claus and C.~Boutilier.
\newblock The dynamics of reinforcement learning in cooperative multiagent
  systems.
\newblock In {\em Proceedings of the Tenth Innovative Applications of
  Artificial Intelligence Conference}, pages 746--752, Madison, Wisconsin, USA,
  July 1998.

\bibitem{Wellman}
J.~Hu and M.~P. Wellman.
\newblock Multiagent reinforcement learning: Theoretical framework and an
  algorithm.
\newblock In {\em Proceedings of the Fifteenth International Conference on
  Machine Learning}, pages 242--250, 1998.

\bibitem{Kaelbling96}
L.~P. Kaelbling, M.~L. Littman, and A.~W. Moore.
\newblock Reinforcement learning: A survey.
\newblock {\em Journal of Artificial Intelligence Research}, 4, 1996.

\bibitem{Littman94}
M.~L. Littman.
\newblock Memoryless policies: Theoretical limitations and practical results.
\newblock In {\em From Animals to Animats 3}, Brighton, UK, 1994.

\bibitem{MeuleauUAI99}
N.~Meuleau, L.~Peshkin, K.-E. Kim, and L.~P. Kaelbling.
\newblock Learning finite-state controllers for partially observable
  environments.
\newblock In {\em Proceedings of the Fifteenth Conference on Uncertainty in
  Artificial Intelligence}, pages 427--436. Morgan Kaufmann, 1999.

\bibitem{Narendra}
K.~S. Narendra and M.~A. Thathachar.
\newblock {\em Learning Automata}.
\newblock Prentice Hall, 1989.

\bibitem{Osborne}
M.~J. Osborne and A.~Rubinstein.
\newblock {\em A Course in Game Theory}.
\newblock The MIT Press, 1994.

\bibitem{Peshkin99}
L.~Peshkin, N.~Meuleau, and L.~P. Kaelbling.
\newblock Learning policies with external memory.
\newblock In I.~Bratko and S.~Dzeroski, editors, {\em Proceedings of the
  Sixteenth International Conference on Machine Learning}, pages 307--314, San
  Francisco, CA, 1999. Morgan Kaufmann.

\bibitem{Puterman94}
M.~Puterman.
\newblock {\em Markov Decision Processes}.
\newblock John Wiley \& Sons, New York, 1994.

\bibitem{Schneider99}
J.~Schneider, W.-K. Wong, A.~Moore, and M.~Riedmiller.
\newblock Distributed value functions.
\newblock In I.~Bratko and S.~Dzeroski, editors, {\em Proceedings of the
  Sixteenth International Conference on Machine Learning}, pages 371--378, San
  Francisco, CA, 1999. Morgan Kaufmann.

\bibitem{Singh94a}
S.~Singh, T.~Jaakkola, and M.~Jordan.
\newblock Learning without state-estimation in partially observable {Markovian}
  decision processes.
\newblock In {\em Machine Learning: Proceedings of the Eleventh International
  Conference}. 1994.

\bibitem{Sondik78}
E.~J. Sondik.
\newblock The optimal control of partially observable {Markov} processes over
  the infinite horizon: Discounted costs.
\newblock {\em Operations Research}, 26(2):282--304, 1978.

\bibitem{Sutton98a}
R.~S. Sutton and A.~G. Barto.
\newblock {\em Reinforcement Learning: An Introduction}.
\newblock The MIT Press, Cambridge, Massachusetts, 1998.

\bibitem{Williams87}
R.~J. Williams.
\newblock A class of gradient-estimating algorithms for reinforcement learning
  in neural networks.
\newblock In {\em Proceedings of the IEEE First International Conference on
  Neural Networks}, San Diego, California, 1987.

\bibitem{Williams92}
R.~J. Williams.
\newblock Simple statistical gradient-following algorithms for connectionist
  reinforcement learning.
\newblock {\em Machine Learning}, 8(3):229--256, 1992.

\end{thebibliography}
\end{document}